# Multistep Speed Prediction on Traffic Networks: A Graph Convolutional Sequence-to-Sequence Learning Approach with Attention Mechanism

Zhengchao Zhang[a], Meng Li[a,b], Xi Lin[b], Yinhai Wang[c,a], Fang He[d,b*]

[a]*Department of Civil Engineering, Tsinghua University, Beijing 100084, P.R. China*

[b]*Tsinghua-Daimler Joint Research Center for Sustainable Transportation, Tsinghua University, Beijing 100084, P.R.China*

[c]*Department of Civil and Environmental Engineering, University of Washington, Seattle, WA 98195, USA*

[d]*Department of Industrial Engineering, Tsinghua University, Beijing 100084, P.R. China*

**Abstract:** Multistep traffic forecasting on road networks is a crucial task in successful intelligent transportation system applications. To capture the complex non-stationary temporal dynamics and spatial dependency in multistep traffic-condition prediction, we propose a novel deep learning framework named attention graph convolutional sequence-to-sequence model (AGC-Seq2Seq). In the proposed deep learning framework, spatial and temporal dependencies are modeled through the Seq2Seq model and graph convolution network separately, and the attention mechanism along with a newly designed training method based on the Seq2Seq architecture is proposed to overcome the difficulty in multistep prediction and further capture the temporal heterogeneity of traffic pattern. We conduct numerical tests to compare AGC-Seq2Seq with other benchmark models using a real-world dataset. The results indicate that our model yields the best prediction performance in terms of various prediction error measures. Furthermore, the variation of spatiotemporal correlation of traffic conditions

[*] Corresponding author. Email: *fanghe@tsinghua.edu.cn*.

under different perdition steps and road segments is revealed through sensitivity analyses.

## 1. INTRODUCTION

Automobile use has significantly increased in the past few decades owing to the steady development in both technology and economy. However, the increased automobile use has resulted in a series of social problems such as traffic congestion, traffic accidents, energy overconsumption, and carbon emissions (Gao et al., 2011). The intelligent transportation system (ITS) has been considered as a promising solution to improve transportation management and services (Qureshi and Abdullah, 2013; Lin et al., 2017). The success of ITS applications relies on accurate and timely traffic status information. The applications with high-precision traffic prediction (e.g., advanced traffic management systems and advanced traveler information systems) not only benefit travelers' route planning and departure time scheduling (Yuan et al., 2011) but also provide insightful information for traffic control to improve traffic efficiency and safety (Belletti et al., 2017). Therefore, short-term traffic flow forecasting has always attracted many scholars' interest (Vlahogianni et al., 2004).

Substantial efforts have been conducted to develop methods for traffic prediction in the literature; however, some major challenges remain. Vlahogianni et al. (2014) provided a comprehensive review of the entire spectrum of the short-term traffic forecasting literature up to 2014 and reported the following potential directions for future research:

- Traffic prediction on traffic networks should be emphasized more.
- Multistep for medium-long term forecasting is more adaptive to practical applications.
- Research on incorporating both temporal characteristics of traffic flow and spatial

dependencies on traffic network still deserves more comprehensive investigation.

Multistep traffic forecasting on road networks is challenging primarily due to the non-Euclidean topology structure and stochastic characteristic of the non-stationary time-varying traffic patterns, and inherent difficulty in multistep prediction. Hence, we propose a novel deep learning structure, named the **a**ttention **g**raph **c**onvolutional **seq**uence-**to-seq**uence model (AGC-Seq2Seq). Specifically, we integrate the graph convolutional network and attention mechanism into a Seq2Seq framework to develop the prediction model that can depict the spatial-temporal correlation in multistep traffic prediction. Furthermore, considering that the existing training method for the Seq2Seq model is not suitable for time-series problems, we hereby design a new training method in our proposed framework. To summarize, the primary contributions of this paper are listed as follows:

i. We propose a novel deep learning framework, named AGC-Seq2Seq, which extracts the features from temporal and spatial domains simultaneously through the Seq2Seq model and graph convolution layer. To overcome the multistep prediction challenge and capture the temporal heterogeneity of urban traffic pattern, the attention mechanism is further incorporated into the model. Validated by the real-world traffic data provided by A-map (Gaode navigation), the proposed model yields a significant improvement over other state-of-the-art benchmarks in terms of various major error measures under different prediction intervals.

ii. We design a new training method for the Seq2Seq framework aiming at multistep traffic prediction to replace the existing ones (e.g., *teacher forcing* and *scheduled sampling*). It coordinates multidimensional features (e.g., historical statistic information and time-of-day) with spatial-temporal speed variables in one end-to-end deep learning structure and enables the input for the testing periods to agree with the training periods.

iii. Based on the proposed model, we explore the variation of spatial and temporal

correlations of traffic conditions under different perdition steps and road segments.

The remainder of this paper is organized as follows. Section 2 first reviews the existing research. Section 3 formulates the short-term traffic speed forecasting problem, and describes the structure and mathematical formulation of the proposed AGC-Seq2Seq model. Section 4 compares the prediction performances of the proposed model with other benchmark models based on the real-world dataset in Beijing and presents sensitive analyses. Finally, Section 5 concludes the paper.

## 2. LITERATURE REVIEW

Traffic flow/condition forecasting has been studied for decades, and various emerging methods are constantly used to model traffic characteristics. With the rapid development of real-time traffic data collection methods, data-driven approaches through enormous historical data to capture similar traffic patterns prevail in recent years. As reported by Li et al. (2017), statistical models, shallow machine learning models and deep learning models are three major representative categories.

Statistical models can predict future values based on previously observed values by time-series analysis. The autoregressive integrated moving average (ARIMA) model (Ahmed and Cook, 1979), Kalman filter (Okutani and Stephanedes, 1984), and their variations (Williams and Hoel, 2003; Guo et al., 2014) are among the most consolidated approaches. However, simple time-series models typically rely on the stationary assumption, which is inconsistent with non-stationary characteristics of urban traffic dynamics. Specifically, for multistep prediction, the posterior predicted values are based on the prior predicted values; thus, the prediction errors could propagate step by step. In this context, it is difficult to satisfy the high-precision requirement using simple time-series models.

Meanwhile, machine learning methods have shown promising capabilities in traffic forecasting studies. The artificial neural network model (Vlahogianni et al., 2005),

Bayesian networks (Fusco et al., 2016), support vector machine model (Castro-Neto et al., 2009), K-nearest neighbors model (Zhang et al., 2013; Habtemichael and Cetin, 2016; Cai et al., 2016) and random forest model (Hamner, 2011) all yield satisfactory results in traffic flow forecasting. However, the performances of machine learning models depend heavily on manually selected features, and well-recognized guidelines to choose the appropriate features are not available in general since the key features are problem-wise. Therefore, using elementary machine learning approaches may not yield the prospective outcomes for complicated prediction tasks.

More recently, deep learning models have been widely and successfully employed in computer science; meanwhile, it has drawn substantial attention in the transportation field. Huang et al. (2014) employed the deep belief network for unsupervised feature learning, which was proven efficient in traffic flow prediction. Lv et al. (2015) applied a stacked auto encoder model to learn generic traffic flow features. Ma et al. (2015) used the long short-term memory neural network (LSTM) to capture nonlinear traffic dynamics effectively. Polson and Sokolov (2017) combined $L_1$ regularization and a multilayer network activated by the tanh function to detect the sharp nonlinearities of traffic flow. However, the models with deep architectures above mainly aim at modeling a single sequence, which fails to reflect spatial correlations on traffic networks.

Meanwhile, convolutional neural networks (CNN) offer an efficient architecture to extract meaningful statistical patterns in large-scale, and high-dimensional datasets. The capability of CNNs in learning local stationary structures resulted in breakthroughs in image and video recognition tasks (Defferrard et al., 2016). In transportation, efforts have been conducted to apply CNN structures to extract spatial correlation on traffic networks. Ma et al. (2017) proposed a deep convolutional neural network for traffic speed prediction, where spatial-temporal traffic dynamics are converted to images. Wang et al. (2017) processed an expressway as a band image, and subsequently proposed the error-feedback recurrent convolutional neural network structure for continuous traffic speed prediction. Ke et al. (2017) partitioned the urban area into uniform grids and subsequently combined a convolutional layer with an LSTM layer to predict the on-demand passenger demand in each grid. All of the aforementioned research converted traffic network to regular grids because the CNNs are restricted to processing Euclidean-structured data. However, the time series on road networks in traffic forecasting are continuous sequences distributed over a topology graph, which is a typical representative of non-Euclidean-structured data (Narang et al., 2013); in this case, the original CNN structure may not be applicable. To fill this gap, the graph convolutional network (GCN) was developed to generalize the convolution on non-Euclidean domains in the context of spectral graph theory (Kipf and Welling, 2016). Several newly published studies conducted graph convolution on traffic prediction. Spectral-based graph convolution was adopted and combined with temporal convolution (Yu et al., 2017) and the recurrent neural network (RNN) (Li et al., 2017) to forecast traffic states. Later, Cui et al. (2018) applied high-order graph convolution to learn the interactions between links on the traffic network. The studies above do not directly define the graph convolution on road networks, but construct the traffic detector graph through computing the pairwise distances between sensors with threshold Gaussian kernel. Moreover, the temporal correlation of traffic conditions is not considered, either.

To summarize, the evolution of traffic conditions on urban networks exhibits spatial and temporal dependencies, substantially. In this paper, we are devoted to proposing a customized deep learning framework, which integrates the attention mechanism and the graph convolutional network into a Seq2Seq model structure, to simultaneously capture the complex non-stationary temporal dynamics and spatial dependency in multistep traffic-condition prediction.

## 3. AGC-SEQ2SEQ DEEP LEARNING FRAMEWORK

### 3.1. Preliminaries

In this subsection, we interpret the definitions and notations of the variables used herein.

*(1) Road network topology*

The road network is modeled as a directed graph $\mathcal{G}(\mathcal{N},\mathcal{L})$ according to the driving direction, where the node set $\mathcal{N}$ represents the intersections (detectors or selected demarcation points on the freeway), and the link set $\mathcal{L}$ represents the road segments, as shown in Figure 1. $\boldsymbol{A}$ is the adjacency matrix of the link set, and the dummy variable $\boldsymbol{A}(i,j)$ denotes whether link $i$ and link $j$ are connected, i.e., $\boldsymbol{A}(i,j) = \begin{cases} 1, & l_i \text{ and } l_j \text{ are connected along driving direction} \\ 0, & \text{otherwise} \end{cases}$.

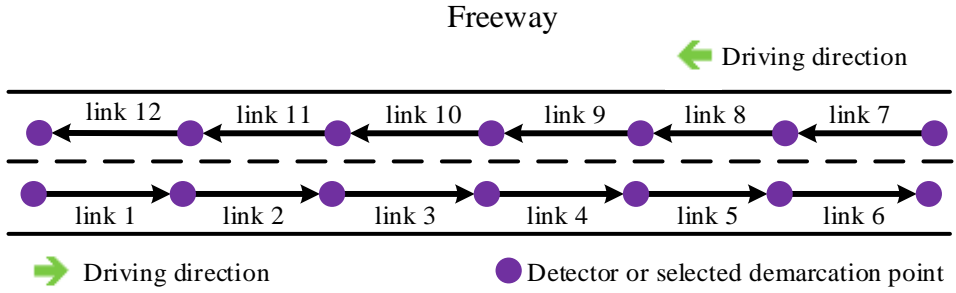

(a) Freeway

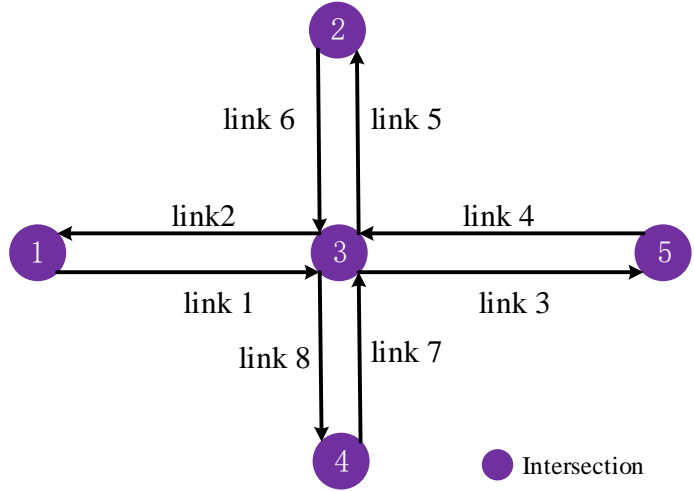

(b) Urban road network

Figure 1. Topology of traffic networks

*(2) Traffic speed*

The speed at the $t^{\text{th}}$ time slot (e.g., 5 min) of road segment $l_i$ ($\forall l_i \in \mathcal{L}$) is defined as the average speed of floating cars during this time interval on the road segment, which is denoted by $v_t^i$. The speed of the road network at the $t^{\text{th}}$ time slot is defined as the vector $V_t \in \mathbb{R}^{|\mathcal{L}|}$ ($|\mathcal{L}|$ is the cardinality of link set $\mathcal{L}$ in the underlying road network), where the $i^{\text{th}}$ element is $(V_t)_i = v_t^i$.

As a classical time-series prediction problem, the nearest m-step observation data can provide valuable information for multistep traffic speed forecasting. In addition to the real-time traffic speed information, some exogenous variables such as the time-of-day, weekday-or-weekend, and historical statistic information are also helpful to predict the future traffic speed. We introduce these variables in the following part.

*(3) Time-of-day and weekday-or-weekend*

Because the speed of each road segment is aggregated as the average value in 5 min, the time-of-day is transformed into an ordered integer $N$, e.g., 00:00-00:05 as $N_t = 1$, and 7:00-7:05 as $N_t = 85$ $(7*12+1)$. The weekday-or-weekend is denoted by the dummy variable $p_t$ that distinguishes different traffic characteristics between weekdays and weekends.

*(4) Historical statistic information*

The daily trend of the traffic status can be captured by introducing historical statistic information into the prediction model. The historical average speed, median speed, maximum speed, minimum speed, and standard deviation at the $t^{\text{th}}$ time slot of road segment $l_i$ are defined as the average value, median value, maximum value, minimum value, and standard deviation in the training dataset, respectively, which are denoted by $v_{t,average}^i$, $v_{t,median}^i$, $v_{t,max}^i$, $v_{t,min}^i$ and $d_t^i$, respectively.

*(5) Problem formulation*

The task of traffic speed prediction is to use the previously observed speed records to forecast the future values of each road segment in a certain period. The multistep traffic speed problem can be formulated as

$$\widehat{V}_{t+n} = \underset{V_{t+n}}{\mathrm{argmax}}\ \mathrm{Pr}(V_{t+n}|V_t, V_{t-1}, \cdots, V_{t-m}; \mathcal{G}) \tag{1}$$

where $\widehat{V}_{t+n}(n = 1,2,3,\cdots)$ represents the $n^{\text{th}}$-step predicted speed of the underlying road network, and $\{V_t, V_{t-1}, \cdots, V_{t-m}\ |\ m = 1,2,\cdots\}$ is the relevant previously observed value vector. $\mathrm{Pr}(\cdot\,|\,\cdot)$ is the conditional probability function.

**3.2 Graph Convolution on Traffic Networks**

Graph convolution extends the applicable scope of standard convolution from regular grids to general graphs by manipulating in the spectral domain. To introduce the general $K$-order graph convolution, we first define the $K$-hop neighborhoods for each road segment $l_i \in \mathcal{L}$ as $\mathcal{H}_i(K) = \{l_j \in \mathcal{L}\,|\,d(l_i, l_j) \leq K\}$ in the context of road network topology (introduced in section 3.1), where $d(l_i, l_j)$ represents the minimum number of needed links among all the walks from $l_i$ to $l_j$.

It is typical that the adjacency matrix is exactly the one-hop neighborhood matrix $A$, and the $K$-hop neighborhood matrix can be acquired by calculating the $K^{\text{th}}$ power of $A$. To imitate the Laplacian matrix, we add the diagonal element to the neighborhood matrix, which is defined as

$$A_{GC}^K = \mathrm{Ci}(A^K + I) \tag{2}$$

where $\mathrm{Ci}(\cdot)$ is a clip function for the matrix by modifying each nonzero element to 1; thus, $A_{GC}^K(i,j) = 1\ for\ l_j \in \mathcal{H}_i(K)\ or\ i = j$; otherwise, $A_{GC}^K(i,j) = 0$. The identity matrix $I$ added to $A^K$ renders the convolution self-accessible in the topology graph.

Based on the abovementioned neighborhood matrix, a concise version of graph convolution (e.g., Cui et al., 2018) can be defined as follows.

$$\boldsymbol{V}_t(K) = (\boldsymbol{W}_{GC} \odot \boldsymbol{A}_{GC}^K) \cdot \boldsymbol{V}_t \tag{3}$$

where $\boldsymbol{W}_{GC}$ is a trainable matrix with the same size of $\boldsymbol{A}$. The operator $\odot$ refers to the Hadamard product that conducts the element-wise multiplication operation. Through the element-wise multiplication, $(\boldsymbol{W}_{GC} \odot \boldsymbol{A}_{GC}^K)$ will produce a new matrix with trainable parameters on the $K$-hop neighbor positions and zero on the remaining positions. Therefore, $(\boldsymbol{W}_{GC} \odot \boldsymbol{A}_{GC}^K) \cdot \boldsymbol{V}_t$ can be understood as spatial discrete convolution for $\boldsymbol{V}_t$. In consequence, $\boldsymbol{V}_t(K)$ is the spatially-fused speed vector at the time $t$. Its $i^{\text{th}}$ element $v_t^i(K)$ represents the spatially-fused speed of the road segment $l_i \in \mathcal{L}$ at the time $t$ that incorporates the information of all the neighbor road segments in $\mathcal{H}_i(K)$.

Further, Equation (3) can be decomposed into a one-dimensional convolution that is flexible and suitable for parallel computing.

$$v_t^i(K) = (\boldsymbol{W}_{GC}[i] \odot \boldsymbol{A}_{GC}^K[i])^T \cdot \boldsymbol{V}_t \tag{4}$$

$\boldsymbol{W}_{GC}[i]$ and $\boldsymbol{A}_{GC}^K[i]$ are the $i^{\text{th}}$ row of $\boldsymbol{W}_{GC}$ and $\boldsymbol{A}_{GC}^K$, respectively. An example of $\boldsymbol{A}_{GC}^K[i]$ on the road network is shown in Figure 2, where the road segment $i$ is in red line and neighbor links are in blue lines.

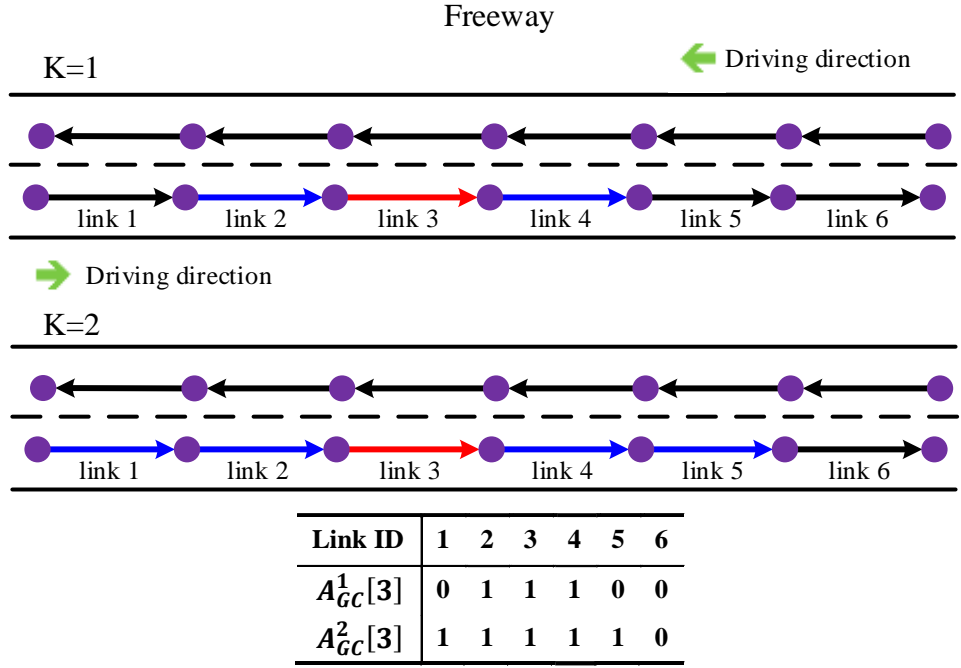

(a) Freeway

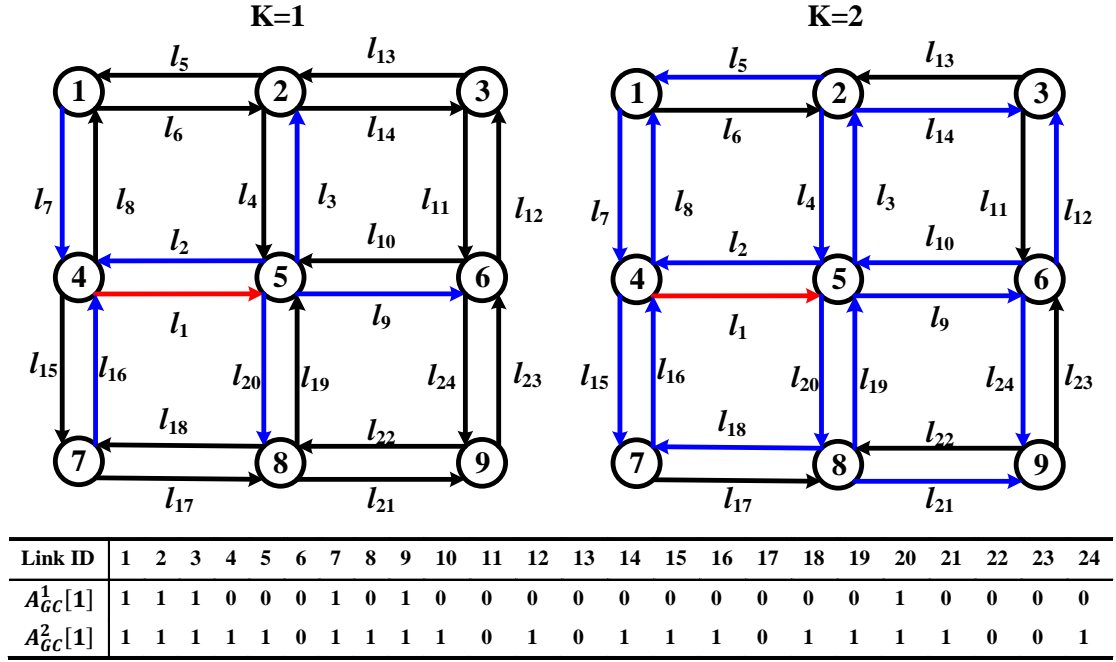

(b) Urban road network

Figure 2. Illustration of $A_{GC}^{K}[i]$

## 3.3 Attention Graph Convolutional Sequence-to-Sequence Model (AGC-Seq2Seq)

In this subsection, we propose the novel AGC-Seq2Seq model that integrates spatial-temporal variables and exogenous information into the deep learning architecture for multistep traffic speed prediction.

To capture time-series characteristics and obtain multistep outputs, we adopt the Seq2Seq model as the basic structure for the whole approach that is composed of two connected RNN modules with independent parameters (Sutskever et al., 2014; Cho et al., 2014). To overcome the fixed output timestamp of the RNN structure, the Seq2Seq model encodes the time-series input in the encoder part to satisfy the temporal dependencies, and the decoder produces the target outputs organized by time steps from the *context vector*. The framework of the proposed AGC-Seq2Seq model is shown in Figure 3. Specifically, the graph convolution operation is first utilized to capture the spatial characteristics. Subsequently, the spatial-temporal variable $v_{t-j}^i(K)$ is fused with exogenous variable $E_{t-j}$ (including the information of time-of-day and weekday-or-weekend) to construct the input vector, which is then fed into the encoder of Seq2Seq model. The procedure above is demonstrated in the following equations.

$$v_{t-j}^i(K) = (W_{GC}[i] \odot A_{GC}^K[i])^T \cdot V_{t-j}, \qquad 0 \leq j \leq m \tag{5}$$

$$E_{t-j} = [N_{t-j}; p_{t-j}] \tag{6}$$

$$X_{t-j}^i = [v_{t-j}^i(K); E_{t-j}] \tag{7}$$

where $N_{t-j}$ and $p_{t-j}$ are defined in section 3.1; and the operator $[\cdot \, ; \, \cdot]$ concatenates two tensors along the same dimensions.

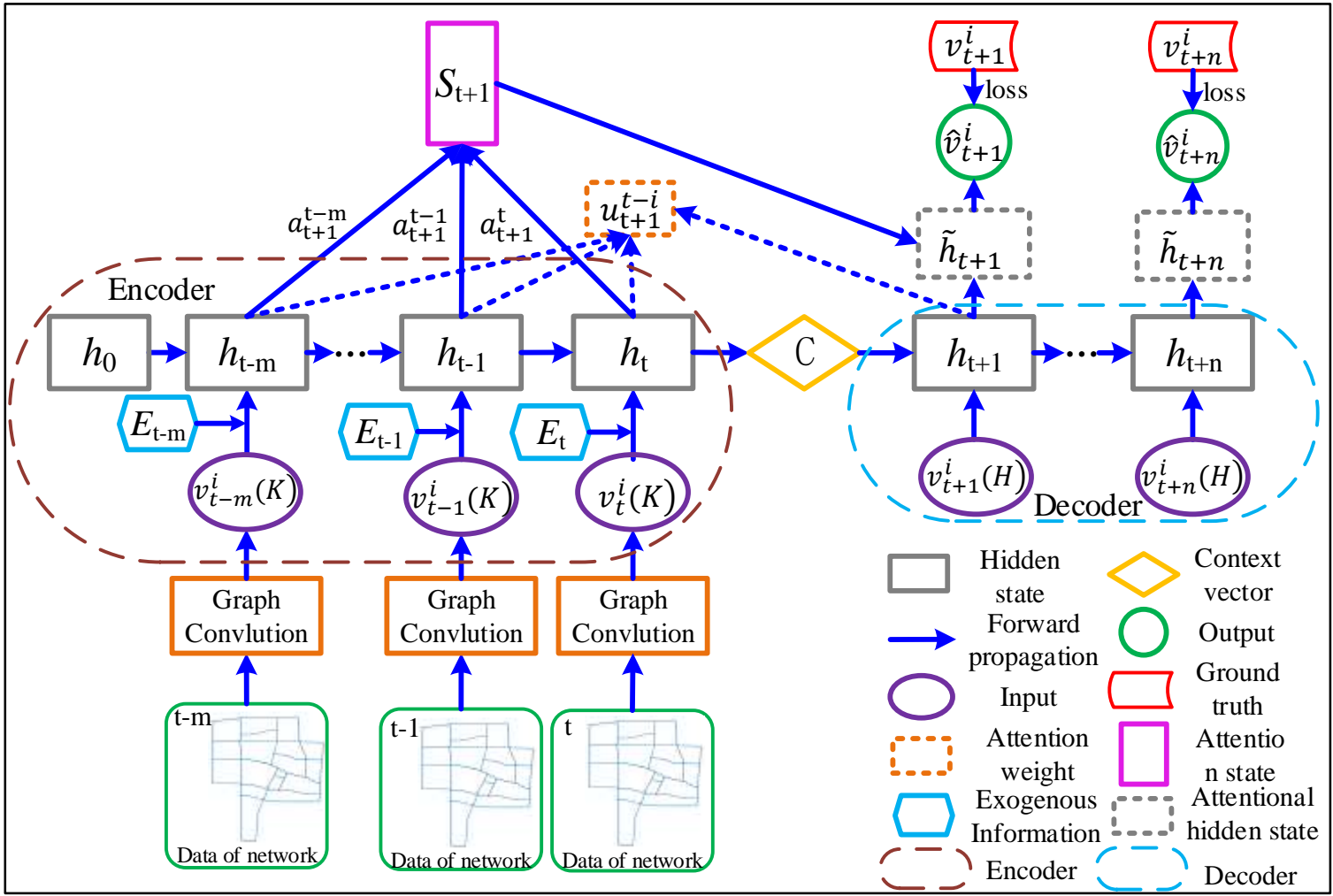

Figure 3. Structure of the proposed model (taking $t+1$ time step attention for example)

Then, in the encoder part, as demonstrated in Equations (8)–(9) below, at the time step $t-j, j \in \{0, \cdots, m\}$, the previous hidden status $h_{t-j-1}$ is passed to the current time stamp together with input $X_{t-j}$ to calculate $h_{t-j}$. Therefore, the context vector $C$ stores all the information of the encoder including the hidden states $(h_{t-m}, h_{t-m+1}, \cdots, h_{t-1})$ and input vector $(X_{t-m}, X_{t-m+1}, \cdots, X_t)$, which is further designed as a connector between encoder and decoder parts.

$$h_{t-j} = \begin{cases} \text{Cell}_{\text{encoder}}(h_0, X_{t-j}), & j = m \\ \text{Cell}_{\text{encoder}}(h_{t-j-1}, X_{t-j}), & j \in \{0, \cdots, m-1\} \end{cases} \qquad (8)$$

$$C = h_t \qquad (9)$$

where $h_0$ is the initial hidden status and typically set as a zero vector; $\text{Cell}_{\text{encoder}}(\cdot)$ is the calculation function for the encoder that is decided by the adopted RNN structure.

In the decoder part, the core idea is leveraging the context vector $C$ as the initial hidden status, and subsequently decoding the output sequence step by step. In consequence, at the time stamp $t+j, j \in \{1, \cdots, n\}$, the hidden state $h_{t+j}$ not only contains the input information, but also considers the previous output status $(h_{t+1}, h_{t+2}, \cdots, h_{t+j-1})$.

The inputs of the decoder are dependent on the training method. *Teacher forcing* is a popular training strategy used in natural language processing. In the teacher-forcing training strategy, the ground truths (target sequence) are fed into the decoder for the training, and at the testing stage, the previously generated predictions are utilized as input for the later time stamp. However, this method is not suitable for the time-series problem primarily because of the discrepant distributions of the decoder inputs between the training and testing periods. Li et al. (2017) mitigated this issue by using *scheduled sampling* that randomly selects either the ground truth or the previous prediction to feed the model with the setting probability $\epsilon$. However, it will inevitably increase the complexity of the model and calculation burden.

To overcome the issues above, we propose a new training method employing the historical statistic information and time-of-day as inputs. In the time-series prediction problem, historical statistic information can be obtained both in the training and testing stages; in this context, the distribution of decoder inputs between the training and testing periods will synchronize with each other, thus solving the dilemma of *teacher forcing*. Moreover, because historical statistic information is critical in multistep forecasting, adding it to the model is expected to enhance the prediction accuracy. Accordingly, the equations below are used to calculate the hidden state in the decoder at the time $t+j$, $j \in \{1, \cdots, n\}$.

$$v_{t+j}^i(H) = [N_{t+j}; v_{t+j,average}^i; v_{t+j,median}^i; v_{t+j,max}^i; v_{t+j,min}^i; d_{t+j}^i] \quad (10)$$

$$h_{t+j} = \begin{cases} \text{Cell}_{\text{decoder}}\left(C, v_{t+j}^i(H)\right), j = 1 \\ \text{Cell}_{\text{decoder}}\left(h_{t+j-1}, v_{t+j}^i(H)\right), j \in \{2, \cdots, n\} \end{cases} \quad (11)$$

where $\text{Cell}_{\text{decoder}}(\cdot)$ is the calculation function for the decoder, which is similar to that of the encoder.

We employ the Gated Recurrent Unit (Chung et al., 2014) as the inner structure for both the encoder and decoder (shown in Figure 4). It demonstrates competitive performance and a simpler structure than the standard LSTM. The calculation procedure of $\text{Cell}_{\text{encoder}}(\cdot)$ and $\text{Cell}_{\text{decoder}}(\cdot)$ is shown in Equations (12)–(17) below.

$$z_t = \sigma(W_z \cdot [h_{t-1}; x_t] + b_z) \quad (12)$$
$$r_t = \sigma(W_r \cdot [h_{t-1}; x_t] + b_r) \quad (13)$$
$$c_t = \tanh(W_c \cdot [r_t \odot h_{t-1}; x_t] + b_c) \quad (14)$$
$$h_t = (1 - z_t) \odot h_{t-1} + z_t \odot c_t \quad (15)$$
$$\sigma(x) = \frac{1}{1 + e^{-x}} \quad (16)$$
$$\tanh(x) = \frac{e^x - e^{-x}}{e^x + e^{-x}} \quad (17)$$

In the equations above, $z_t$ and $r_t$ are the update gate and the reset gate, respectively.

$c_t$ is the candidate output. $\sigma(\cdot)$ and $\tanh(\cdot)$ are the two widely used nonlinear activation functions that map the input into $(0,1)$ and $(-1,1)$, respectively. $W_z$, $W_r$, and $W_c$ are the weight matrices that achieve the fully connected layer, while $b_z, b_r, b_c$ are the corresponding bias vectors.

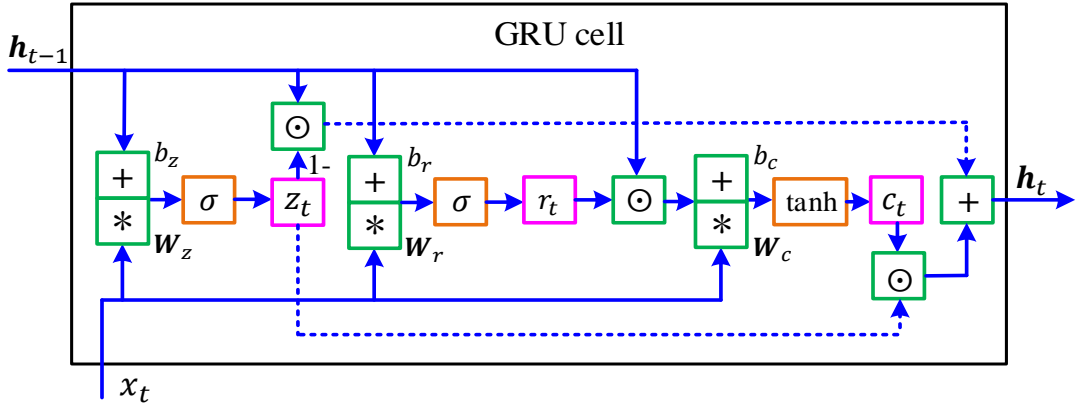

Figure 4. Sketch of GRU

To capture the temporal heterogeneity of traffic pattern, we further integrate the attention mechanism (Bahdanau et al., 2014; Luong et al., 2015) into the model. The key concept of the attention mechanism is adding the attention vector for each time step that captures the relevance of the source-side information to help predict the traffic speed. At the time step $t+j$, $j \in \{1, \cdots, n\}$, the attention function, defined by Equations (18)-(20), maps query $h_{t+j}$ and a set of keys $(h_{t-m}, \cdots, h_{t-1}, h_t)$ to the attention vector $S_{t+j}$. As given by Equations (18)–(20) below, $S_{t+j}$ is computed as a weighed sum of the keys, where the weight assigned to each key is obtained by a compatibility function of the query with the corresponding key.

$$u_{t+j}^{t-i} = q^T \tanh(h_{t+j} W_f h_{t-i}), i = 0, 1, \cdots, m \tag{18}$$

$$a_{t+j}^{t-i} = \text{softmax}(u_{t+j}^{t-i}) = \frac{\exp(u_{t+j}^{t-i})}{\sum_{r=1}^{m} \exp(u_{t+j}^{t-r})}, i = 0, 1, \cdots, m \tag{19}$$

$$S_{t+j} = \sum_{i=1}^{m} a_{t+j}^{t-i} h_{t-i} \tag{20}$$

where $u_{t+j}^{t-i}$ can be used construed as to measure the similarity between $\boldsymbol{h}_{t+j}$ and $\boldsymbol{h}_{t-i}$, calculated by Equation (18), and in this study, we employ the *Luong Attention* form (Luong et al., 2015) as the compatibility function with trainable weight matrix $\boldsymbol{W}_f$ and vector $\boldsymbol{q}^T$ to adjust the dimension of the result; $a_{t+j}^{t-i}$ is the normalization of $u_{t+j}^{t-i}$ and is further used as the weight coefficient with the corresponding encoder hidden state $\boldsymbol{h}_{t-i}$ to calculate $\boldsymbol{S}_{t+j}$.

As shown in Figure 3, the attentional hidden state $\widetilde{\boldsymbol{h}}_{t+j}$ is composed of the attention vector $\boldsymbol{S}_{t+j}$ and original hidden state $\boldsymbol{h}_{t+j}$ through a simple concatenation, as shown in Equation (21). Equation (22) denotes the linear transformation from the hidden state to the output. The dimensions of the weighted parameter matrix $\boldsymbol{W}_v$ and intercept parameter $b_v$ are consistent with the output.

$$\widetilde{\boldsymbol{h}}_{t+j} = \tanh(\boldsymbol{W}_h \cdot [\boldsymbol{S}_{t+k}; \boldsymbol{h}_{t+j}]) \tag{21}$$

$$\hat{v}_{t+j} = \boldsymbol{W}_v \widetilde{\boldsymbol{h}}_{t+j} + b_v \tag{22}$$

To jointly reduce the predictive errors in multiple step prediction, we define the loss as the mean absolute error between $(\hat{v}_{t+1}, \hat{v}_{t+2}, \cdots . \hat{v}_{t+n})$ and $(v_{t+1}, v_{t+2}, \cdots . v_{t+n})$, which is given by

$$loss = \frac{1}{n}\sum_{j=1}^{n}|\hat{v}_{t+j}^i - v_{t+j}^i| \tag{23}$$

All the parameters are updated by minimizing the loss function through the mini-batch gradient descent algorithm in the training stage. A detailed discussion regarding why the Seq2Seq framework is suitable for multistep prediction is presented in the appendix.

## 4 NUMERICAL EXAMPLES

### 4.1. Dataset

The datasets utilized in this study were collected from the users of A-map, which is a smartphone-based navigation application with the most active users in China (Sohu, 2018). The studied site is selected as the entire $2^{nd}$ ring road, which is the most congested among the ring roads in Beijing. As shown in Figure 5(a), we partition the 33KM-in-length $2^{nd}$ ring road into 163 road segments with 200m in length. Furthermore, we calculate the 5-min average speed for each link using the collected trajectory points of anonymous users. The plots of the traffic speed in the $2^{nd}$ ring road on weekdays and weekends are shown in Figure 5(b)-(c) with the x-axis for the longitude, y-axis for the latitude, z-axis for the time and color map for speed.

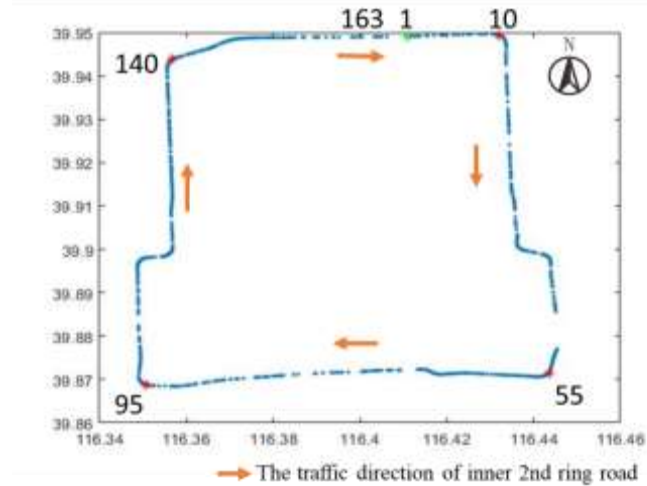

(a) Sketch of road segments in GIS map

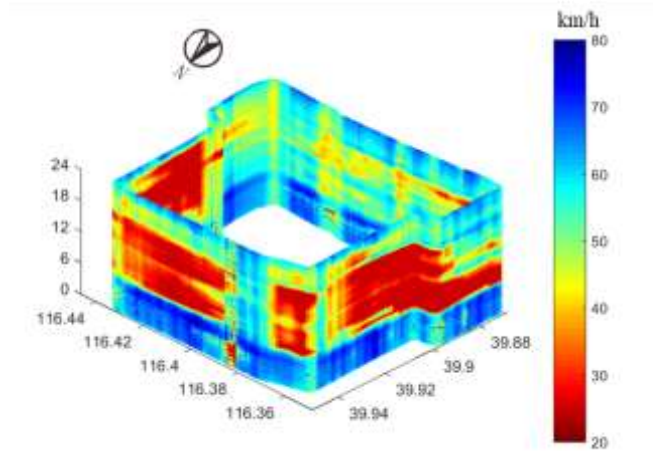

(b) Traffic speed of 2nd ring road on weekdays in July 2016

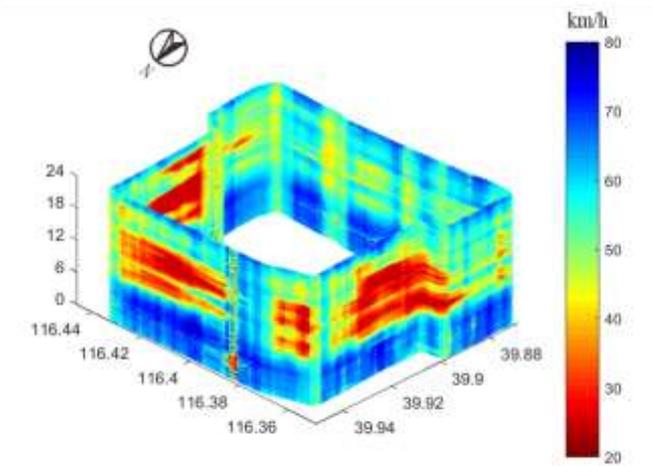

(c) Traffic speed of 2nd ring road on weekends in July 2016

Figure 5. Results of data preprocess

We extract the data from October 1, 2016 to November 30, 2016 for experimental purposes. The extracted dataset is divided into the training set comprised of records between October 1 and November 20, and the testing dataset consisting of the remaining observations between November 21 and November 27. The prediction time horizon is set as 06:00-22:00; thus, every road segment contains 192 data points per day. Figure 6 shows the split of the dataset and the corresponding data size for training and testing. In addition, after the data cleaning procedure, the missing values are filled by the linear interpolation method.

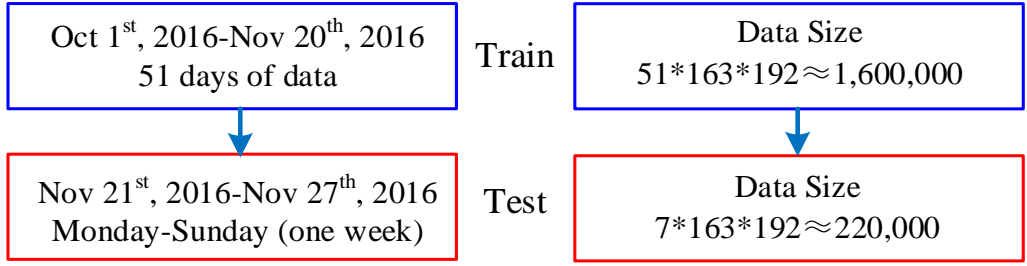

Figure 6. Split of dataset

**4.2 Model comparisons**

In this subsection, the proposed model is compared with other benchmark models, including the traditional time-series analysis approaches (i.e., HA and ARIMA) and state-of-the-art machine learning (i.e., ANN, KNN, SVR, and XGBOOST) /deep learning models (i.e., LSTM, GCN, and Seq2Seq-Att).

- HA: The historical average model predicts the future speed in the testing dataset based on the empirical statistics in the training set, i.e., $v_{t,average}^i$. For example, the average speed during 8:00−8:05 of road segment $l_i \in \mathcal{L}$ is estimated by the mean of all historical speeds in the training dataset during 8:00−8:05 of the same link.

- ARIMA: For the autoregressive integrated moving average $(p, d, q)$ model (Box and Pierce, 1970), the degree of differencing is set as $d = 1$, and the order of autoregressive part and moving average part $(p, q)$ are determined through computing the corresponding Akaike information criterion of the training dataset with $p \in [0,2], q \in [7,12]$.

- ANN: We establish a three-layer artificial neural network (Rumelhart et al., 1988) activated by the sigmoid function, and set the number of hidden neurons twice the dimension of the feature vector. Because the ANN does not differentiate variables across time steps, it fails to capture the temporal dependencies.

- KNN: K-nearest neighbor (Denoeux, 1995) is a lazy learning algorithm that obtains the K-most similar observations in the training set through the Euclidean distance between feature vectors. The predicted value is calculated through the weighted summation of the corresponding future values belonging to the selected observations. The hyper parameter $K$ is chosen through cross validation from 5 to 25.

- SVR: In support vector regression (Suykens and Vandewalle, 1999), the fitting curve is calculated through the mapping feature vectors into the high-dimensional space aided by the kernel function. The kernel function and hyper parameters in the model are selected through cross validation.
- XGBOOST: XGBOOST (Chen and Guestrin, 2016) yields outstanding performance in a broad range of machine learning tasks; it is a scalable end-to-end boosting system based on the tree structure. All the features are reshaped to a vector and fed into XGBOOST for training.
- LSTM: In LSTM (Hochreiter and Schmidhuber, 1997), all the features of each road segment are reshaped to a matrix with one axis as the time steps, and the other axis as the feature category. LSTM takes temporal dependencies into account, but does not capture spatial dependencies.
- GCN: In GCN, the features of all road segments in the underlying traffic network are reshaped to a matrix with one axis as each road segment, and the other axis as the feature category. GCN generalizes the convolution to non-Euclidean domains by the Laplacian matrix of the graph; therefore, it considers spatial correlation, but does not capture temporal dependencies.
- Seq2Seq-Att: In Seq2Seq-Att, the attention mechanism based on the Seq2Seq structure is utilized for traffic prediction along with the new proposed training method. The only difference between the Seq2Seq-Att and AGC-Seq2Seq models is the graph convolution layer.

To ensure fairness, the aforementioned benchmark prediction models have the same input features (the same category and look-back time windows) as those of the AGC-Seq2Seq model, while the traditional time-series model utilizes the whole time-series of speed records in the training set. We consider the look-back time windows as 12 (i.e., $m = 11$), implying that the speed records in the past hour are adopted to predict the future value. The designed 19-dimensional feature vector containing speed observations in the past hour, time-of-day, weekday-or-weekend, and historical statistic information is shown in Table 1.

Table 1. Illustration of feature vector

| No. | Notation | No. | Notation | No. | Notation |
|---|---|---|---|---|---|
| f0 | $N_{t+n}$ | f1 | $v^i_{t+n,average}$ | f2 | $v^i_{t+n,median}$ |
| f3 | $d^i_{t+n}$ | f4 | $v^i_{t+n,max}$ | f5 | $v^i_{t+n,min}$ |
| f6 | $p_{t+n}$ | f7 | $v^i_{t-11}$ | f8 | $v^i_{t-10}$ |
| f9 | $v^i_{t-9}$ | f10 | $v^i_{t-8}$ | f11 | $v^i_{t-7}$ |
| f12 | $v^i_{t-6}$ | f13 | $v^i_{t-5}$ | f14 | $v^i_{t-4}$ |
| f15 | $v^i_{t-3}$ | f16 | $v^i_{t-2}$ | f17 | $v^i_{t-1}$ |
| f18 | $v^i_t$ | | | | |

All the notations are defined in section 3.1. $n$ is fixed according to the prediction step. We evaluate the models via three classical error indexes: mean absolute percentage error (MAPE), mean absolute error (MAE), and root mean squared error (RMSE), given by $\text{MAPE} = \frac{1}{Q}\sum_{i=1}^{Q}\frac{|v_i - \hat{v}_i|}{v_i}$, $\text{MAE} = \frac{1}{Q}\sum_{i=1}^{Q}|v_i - \hat{v}_i|$, and $\text{RMSE} = \sqrt{\frac{1}{Q}\sum_{i=1}^{Q}(v_i - \hat{v}_i)^2}$, where $v_i$ and $\hat{v}_i$ are the $i^{\text{th}}$ ground truth and prediction values of the traffic speed, respectively; $Q$ is the size of the testing dataset.

Our experimental platform is on the server with two CPUs (Intel(R) Xeon(R) CPU E5-2673 v3 @2.40Ghz, 24 cores), 256-GB RAM, and four GPUs (NVIDIA Quadro P5000, 16 GB memory). All the algorithms are coded in the parallel computation structure.

Table 2 shows the comparison of the proposed model and benchmark algorithms for 5 min, 15 min, and 30 min ahead forecasting on the testing dataset. The following phenomena can be observed from the experimental results.

i. AGC-Seq2Seq model outperforms the other benchmarks in terms of all the metrics under all prediction intervals.

ii. The performance of HA is invariant to the increases in the forecasting horizon because it depends only on the historical data.

iii. The performances of all the models under the 5-min forecasting horizons are similar because the traffic status is relatively stable within 5 min.

iv. The deep-learning approaches yield better predictive performances but longer computational time than the traditional machine-learning models.

v. The GCN (which models spatial correlations) outperforms LSTM (which captures the temporal characteristics), providing verification that the consideration of spatial correlations is important in traffic speed forecasting.

vi. The AGC-Seq2Seq model exhibits a distinct improvement over the GCN and Seq2Seq-Att; this emphasizes the importance of capturing the spatial-temporal characteristics simultaneously for the traffic speed forecasting. The running time of the AGC-Seq2Seq model is only slightly higher than those of the GCN and Seq2Seq-Att; it primarily benefits from the advanced parallel computation technology in the GPU module.

Table 2. Prediction performance comparison[1]

(a) 5-min prediction horizon (one step)

| Model | MAPE | MAE | RMSE | Time (s) |
|---|---|---|---|---|
| HA | 30.32% | 7.89 | 10.39 | / |
| ARIMA | 10.65% | 3.64 | 5.40 | 686 |
| ANN | 10.69% | 3.54 | 5.17 | 71 |
| XGBOOST | 10.38% | 3.41 | 5.02 | 58 |
| KNN | 12.26% | 3.88 | 5.84 | 50 |
| SVR | 11.54% | 3.74 | 5.26 | 127 |
| LSTM | 10.25% | 3.48 | 5.23 | 219 |
| GCN[a] | 10.06% | 3.39 | 5.01 | 263 |
| Seq2Seq-Att | 10.10% | 3.40 | 5.11 | 213 |
| AGC-Seq2Seq[a] | **9.57%** | **3.25** | **4.85** | 280 |

(b) 15-min prediction horizon (three steps)

| Model | MAPE | MAE | RMSE | Time(s) |
|---|---|---|---|---|
| HA | 30.32% | 7.89 | 10.39 | / |
| ARIMA | 16.71% | 5.31 | 8.25 | 698 |
| ANN | 16.45% | 4.94 | 7.45 | 72 |
| XGBOOST | 16.07% | 4.82 | 7.34 | 62 |
| KNN | 16.83% | 5.01 | 7.66 | 53 |
| SVR | 16.99% | 5.14 | 7.61 | 125 |
| LSTM | 16.17% | 5.01 | 7.99 | 279 |
| GCN[a] | 14.99% | 4.62 | 7.32 | 363 |
| Seq2Seq-Att | 15% | 4.62 | 7.38 | 350 |
| AGC-Seq2Seq[a] | **14.46%** | **4.47** | **7.12** | 390 |

(c) 30-min prediction horizon (six steps)

| Model | MAPE | MAE | RMSE | Time(s) |
|---|---|---|---|---|
| HA | 30.32% | 7.89 | 10.39 | / |
| ARIMA | 22.82% | 6.99 | 10.6 | 701 |
| ANN | 20.55% | 5.91 | 8.67 | 72 |
| XGBOOST | 20.98% | 5.78 | 8.68 | 63 |
| KNN | 20.05% | 5.79 | 8.71 | 55 |
| SVR | 21.02% | 6.08 | 8.86 | 157 |
| LSTM | 20.70% | 6.40 | 10.03 | 277 |
| GCN[a] | 18.64% | 5.54 | 8.81 | 510 |
| Seq2Seq-Att | 18.4% | 5.36 | 8.64 | 520 |
| AGC-Seq2Seq[a] | **17.94%** | **5.25** | **8.36** | 600 |

---

[1]The order of graph convolution in this experiment is set as one. The high-order situation will be discussed in section 4.3.

We select ARIMA, XGBOOST, and LSTM as the representatives for the time-series models, machine learning models, and deep learning approaches, respectively, to compare their performances with AGC-Seq2Seq under the 5–30-min prediction intervals, as shown in Figure 7. AGC-Seq2Seq tends to demonstrate better performance than other models with the increase in the prediction horizon. Additionally, the ARIMA model performs the worst because of the step-by-step error accumulation in the multistep forecasting scenario.

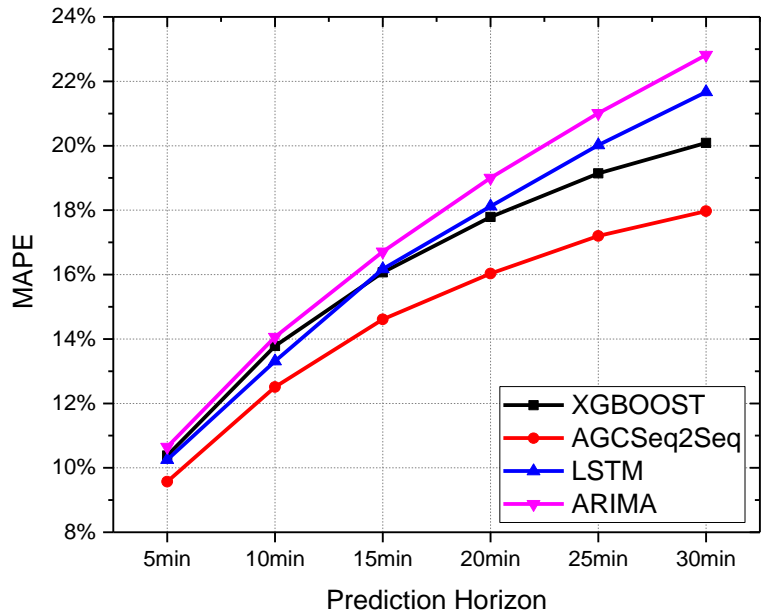

Figure 7. Performance of selected models with varying prediction intervals

Figure 8 shows the prediction values of XGBOOST and AGC-Seq2Seq on the morning peak hours of Nov 23$^{rd}$ (link 29) under the 15-min horizon. It is obvious that the prediction values of XGBOOST lags behind the ground truth when the traffic status oscillates seriously while AGC-Seq2Seq alleviates such problem.

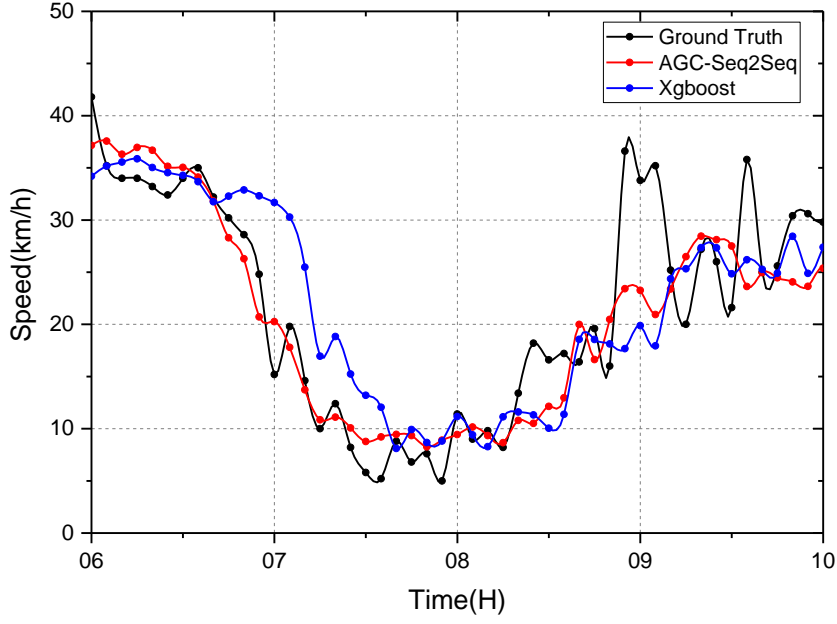

Figure 8. Prediction of morning peak hours (link 29) on Nov 23$^{rd}$

## 4.3 Sensitivity analyses

*(1) Feature importance*

Figure 9 shows the F score (the number of times a feature is used to split the data across all trees, and a higher score indicates the corresponding feature being more import) of the feature vector (shown in Table 1) in the XGBOOST model under a 15-min prediction horizon, which is used widely to assess the importance of the features (Ke et al., 2017). To evaluate the trend of feature importance under different prediction intervals, we divide the feature vector into two major categories: 1) speed records T1 in the past hour (f7-f18) and exogenous information T2 (f0-f6). The F score of feature $fi, i = 0,1,\cdots,18$ is denoted as $F(fi)$. The relative importance of T1 and T2 can be calculated as $\frac{\sum_{i=7}^{18} F(fi)}{\sum_{i=0}^{18} F(fi)}$ and $\frac{\sum_{i=0}^{6} F(fi)}{\sum_{i=0}^{18} F(fi)}$, respectively. Figure 10 shows the relative importance of T1 and T2 under different prediction intervals. The results indicate that the exogenous variables are more important in the long-term prediction than in the short-term prediction. This is because the value of look-back observations degrades gradually with the increase in the forecasting interval.

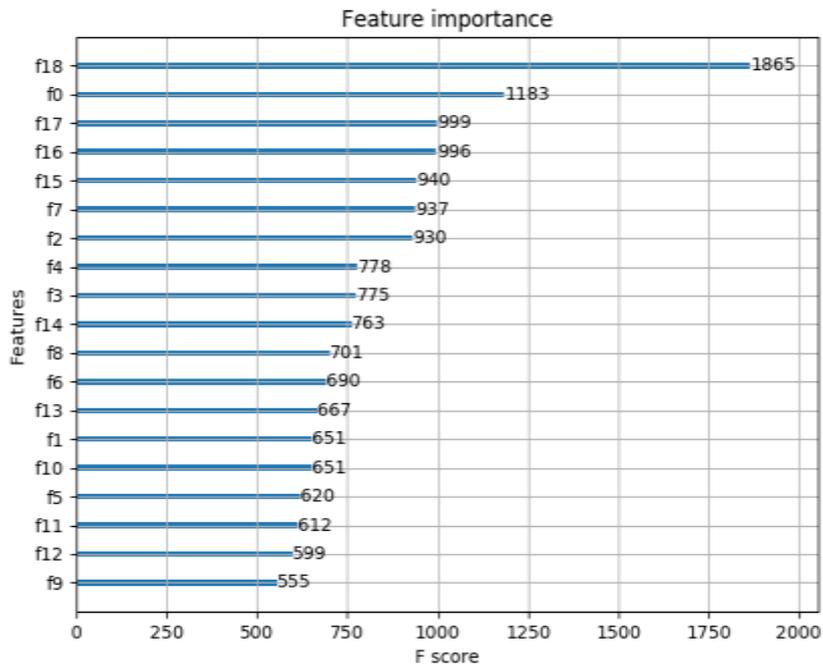

Figure 9. F score of feature vector in XGBOOST model under 15-min horizon

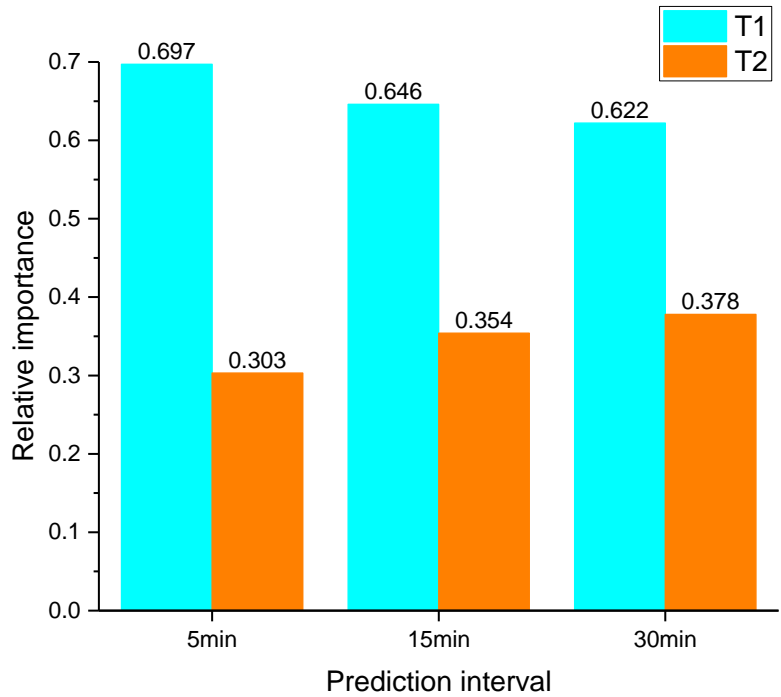

Figure 10. Relative importance of feature under different prediction intervals

*(2) Effect of spatial features in multistep prediction*

Figure 11 shows the curves of prediction error varying with k-hop neighbors in the AGC-Seq2Seq model under 5-min and 15-min time intervals ($k = 0$ is just the case of Seq2Seq-Att model). The slopes of the curves in Figure 11(b) are smaller than those in Figure 11(a), thereby indicating that the effect of increasing spatial information on error reduction is compromised with the increase in the forecasting horizon.

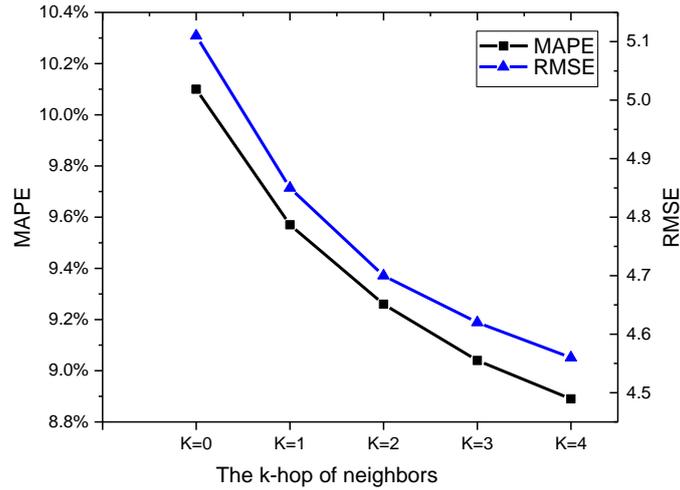

(a) 5-min prediction horizon

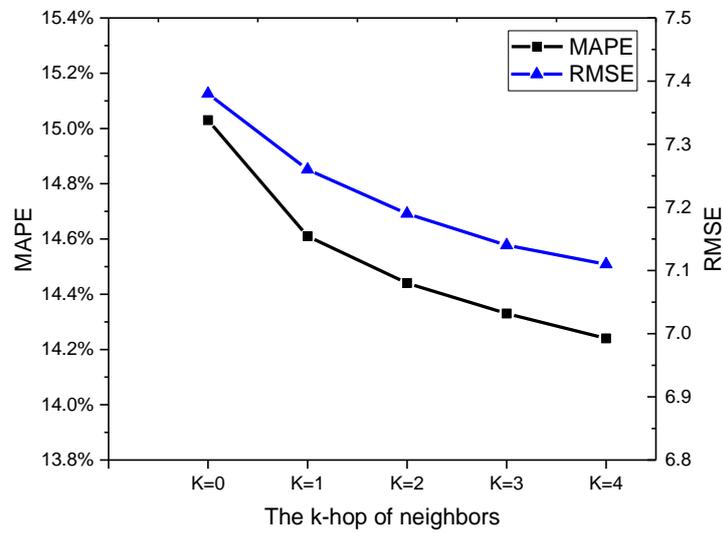

(b) 15-min prediction horizon

Figure 11. Curves of prediction error varying with k-hop neighbors

*(3) Relevance between temporal traffic pattern and attention coefficient*

In the attention mechanism, the coefficient $a_{t+j}^{t-i}$ provides a criterion to measure the relevance between the target and source-side information. Figure 12 visualizes the attention coefficients under two typical scenarios. The attention heatmap of the road segment with drastic status changes (Scenario I) exhibits high values on the look-back observations within the past 15 min, while the corresponding attention coefficients of the smoothly changed traffic status (Scenario II) distribute uniformly among the temporal dimension. This indicates that the prediction model tends to rely on the recent information (within past 15 min) when the traffic status oscillates severely.

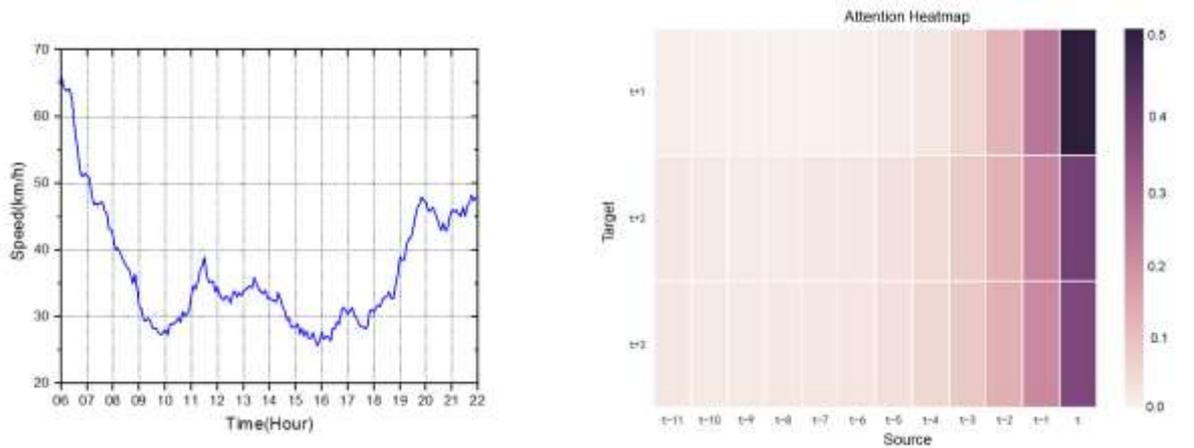

Traffic speed of link 6 (northern 2$^{nd}$ ring road)      Attention heatmap of link 6 under 15-min horizon

(a) Scenario I: Traffic status with drastic changes

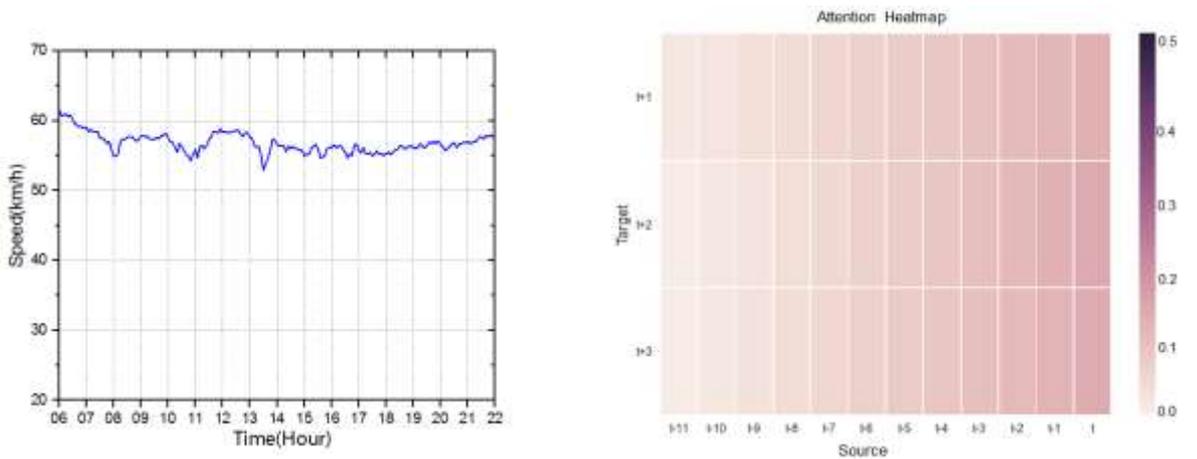

Traffic speed of link 57 (southern 2$^{nd}$ ring road)      Attention heatmap of link 57 under 15-min horizon

(b) Scenario II: Traffic status with smooth changes

Figure 12. Visualization of attention coefficient matrix

# 5 CONCLUSIONS

To tackle the challenge of multistep traffic speed prediction, we are devoted to proposing a sophisticated deep learning approach, i.e., the attention graph convolutional sequence-to-sequence model (AGC-Seq2Seq). The Seq2Seq architecture and graph convolutional operators are combined to learn the spatial-temporal dependencies on traffic networks. The attention mechanism is integrated into the model to capture the temporal heterogeneity of traffic patterns, and the entire architecture is trained with a newly designed method. To validate the effectiveness of the proposed model, we compare it with several benchmark models including the HA, ARIMA, XGBOOST, ANN, LSTM, SVR, KNN, GCN, and Seq2Seq-Att, based on the real-world data of the $2^{nd}$ ring road in Beijing. The results indicate that the proposed model outperforms the benchmark models in terms of the RMSE, MAE, and MAPE under different prediction intervals. Based on the proposed model, we further explore the feature importance, effect of spatial information on multistep prediction, and relevance between traffic temporal pattern and attention coefficients. The evidence from the experiment implies that both the relative importance of the features regarding the speed records in the past hour and the effect of increasing spatial information degrade with the increase in the prediction intervals; for the road segments whose traffic condition changes rapidly, the corresponding attention coefficients take high values for the look-back observations within the past 15 min.

Future studies could include experiments on large urban road networks and further integrating the traffic flow theories into the prediction model, e.g., utilizing the propagation waves of traffic flow to determine the spatial neighbors in a more sophisticated model. From the application perspective, the proposed framework can be integrated with advanced transportation management systems, e.g., providing system-level real-time routing services to reduce peak-hour congestions.

**ACKOWNLEDGMENTS**

This research is supported partially by grants from National Natural Science Foundation of China (71871126, 51622807, U1766205) and partially by Tsinghua-Daimler Joint Research Center for Sustainable Transportation. The authors are grateful to A-map (https://www.amap.com/) for providing their anonymous users' GPS trajectory data.

ACM, 2016:785-794.

Chung, J., Gulcehre, C., Cho, K.H., Bengio, Y., 2014. Empirical Evaluation of Gated Recurrent Neural Networks on Sequence Modeling. Eprint arXiv.

Cui, Z., Henrickson, K., Ke, R., Wang, Y., 2018. High-Order Graph Convolutional Recurrent Neural Network: A Deep Learning Framework for Network-Scale Traffic Learning and Forecasting. Eprint arXiv.

Defferrard, M., Bresson, X., Vandergheynst, P., 2016. Convolutional Neural Networks on Graphs with Fast Localized Spectral Filtering. Eprint arXiv.

Denoeux, T., 1995. A k-nearest neighbor classification rule based on Dempster-Shafer theory. Systems Man & Cybernetics IEEE Transactions on 25(5), 804-813.

Fusco, G., Colombaroni, C., Isaenko, N., 2016. Short-term speed predictions exploiting big data on large urban road networks. Transportation Research Part C 73, 183-201.

Gao, Y., Sun, S., Shi, D., 2011. Network-Scale Traffic Modeling and Forecasting with Graphical Lasso, International Conference on Advances in Neural Networks, pp. 151-158.

Guo, J., Huang, W., Williams, B.M., 2014. Adaptive Kalman filter approach for stochastic short-term traffic flow rate prediction and uncertainty quantification. Transportation Research Part C Emerging Technologies 43, 50-64.

Habtemichael, F.G., Cetin, M., 2016. Short-term traffic flow rate forecasting based on identifying similar traffic patterns. Transportation Research Part C 66, 61-78.

Huang, W., Song, G., Hong, H., Xie, K., 2014. Deep Architecture for Traffic Flow Prediction: Deep Belief Networks With Multitask Learning. IEEE Transactions on Intelligent Transportation Systems 15(5), 2191-2201.

Ke, J., Zheng, H., Yang, H., Xiqun, Chen, 2017. Short-Term Forecasting of Passenger Demand under On-Demand Ride Services: A Spatio-Temporal Deep Learning Approach. Transportation Research Part C Emerging Technologies 85, 591-608.

Kipf, T.N., Welling, M., 2016. Semi-Supervised Classification with Graph Convolutional Networks. Eprint arXiv.

Kuznetsov, V., Mariet, Z., 2018. Foundations of Sequence-to-Sequence Modeling for
32

# APPENDIX

The well-known approaches for multistep prediction could be divided into three categories.

**Method 1**

The intuitive way to make multistep prediction is rolling prediction step by step as given by Equations A1. Time-series models (e.g. MA and ARIMA) utilize this way to make multistep prediction. Since the posterior predicted values are based on prior predicted ones, the error will accumulate through this process.

$$\begin{aligned}
\hat{v}_{t+1} &= \underset{v_{t+1}}{\mathrm{argmax}}\ \Pr(v_{t+1}|v_t, v_{t-1}, \cdots, v_{t-m}) \\
\hat{v}_{t+2} &= \underset{v_{t+2}}{\mathrm{argmax}}\ \Pr(v_{t+2}|\hat{v}_{t+1}, v_t, v_{t-1}, \cdots, v_{t-m+1}) \\
&\vdots \\
\hat{v}_{t+n} &= \underset{v_{t+n}}{\mathrm{argmax}}\ \Pr(v_{t+n}|\hat{v}_{t+n-1}, \hat{v}_{t+n-2}, \cdots, \hat{v}_{t+1}, v_t, v_{t-1}, \cdots, v_{t-m+n-1})
\end{aligned} \quad (A1)$$

**Method 2**

The second method makes multistep prediction by directly learning the dependency of $n^{th}$ prediction value and look-back observations as given by Equations A2. Some classical machine learning models (e.g. SVR, XGBOOST and KNN) utilize this way to make multistep prediction. Since the temporal relevance weakens with the increase of prediction intervals.

$$\hat{v}_{t+n} = \underset{v_{t+n}}{\mathrm{argmax}}\ \Pr(v_{t+n}|v_t, v_{t-1}, \cdots, v_{t-m}) \quad (A2)$$

**Method 3**

The third method makes multistep prediction by training the parameters to cooperatively reduce errors from $1^{st}$ to $n^{th}$ prediction values given by Equations (A3). Kuznetsov and Mariet (2018) provide theoretical proof for the advantage of Sequence-to-Sequence modeling time series problems using this way.

$$\hat{v}_{t+1}, \cdots, \hat{v}_{t+n} = \underset{v_{t+n},\cdots,v_{t+1}}{\mathrm{argmax}} \ \mathrm{Pr}(v_{t+n}, \cdots, v_{t+1} | v_t, v_{t-1}, \cdots, v_{t-m}) \tag{A3}$$